\documentclass{article}

\usepackage{float}
\usepackage{bm}
\usepackage{mathtools}
\usepackage{graphicx}
\usepackage{mdframed}
\usepackage{alltt}
\usepackage{amsmath}
\usepackage{amsfonts}  %
\usepackage{bm}
\usepackage{tcolorbox}
\usepackage{xcolor}
\usepackage{cite}
\usepackage{tabularx}
\usepackage{booktabs}
\usepackage{makecell}
\usepackage[margin=1in]{geometry}
\usepackage{url}

\DeclareMathOperator*{\argmin}{arg\,min}

\usepackage[numbers,square,sort&compress]{natbib}
\begin{document}

\title{Cost-Effective Hallucination Detection for LLMs}

\author{Simon Valentin$^1$\footnotemark[1]
        \and Jinmiao Fu$^2$\footnotemark[1]
        \and Gianluca Detommaso$^3$\thanks{These authors contributed equally to this research.}
        \and Shaoyuan Xu$^2$
        \and Giovanni Zappella$^1$
        \and Bryan Wang$^2$}

\date{
    $^1$Amazon Web Services, Berlin, Germany \\
    \texttt{\{simval,zappella\}@amazon.de}\\
    $^2$Amazon, Seattle, USA \\
    \texttt{\{jinmiaof,shaoyux,brywan\}@amazon.com}\\
    $^3$Helsing, Berlin, Germany \\
    \texttt{detommaso.gianluca@gmail.com}\\
}

\maketitle

\begin{abstract}
Large language models (LLMs)
can be prone to hallucinations --- generating unreliable outputs that are unfaithful to their inputs, external facts or internally inconsistent.
In this work, we address several challenges for post-hoc hallucination detection in production settings.
Our pipeline for hallucination detection entails: first, producing a confidence score representing the likelihood that a generated answer is a hallucination; second, calibrating the score conditional on attributes of the inputs and candidate response; finally, performing detection by thresholding the calibrated score. We benchmark a variety of state-of-the-art scoring methods on different datasets, encompassing question answering, fact checking, and summarization tasks. We employ diverse LLMs to ensure a comprehensive assessment of performance.
We show that calibrating individual scoring methods is critical for ensuring risk-aware downstream decision making.
Based on findings that no individual score performs best in all situations, we propose a multi-scoring framework, which combines different scores and achieves top performance across all datasets.
We further introduce cost-effective
multi-scoring,
which can match or even outperform more expensive detection methods, while significantly reducing computational overhead.
\end{abstract}

\section{Introduction}
Despite their impressive capabilities,  large language models (LLMs) can be prone to generating hallucinations --- undesirable outputs that are incorrect, unfaithful, or inconsistent with respect to the inputs (or the output itself)~\cite{zhao2023survey}.
These unreliable behaviors pose significant risks for adopting LLMs in real-world applications.
Challenges in detecting hallucinations lie, among other things, in hallucinations taking different forms, being context-dependent and sometimes being in conflict with other desirable properties of generated text~\cite{ji2023survey, zhang2023siren}.
Hallucinations may be harmless in some contexts, but can be undesired or potentially dangerous in other applications (e.g., erroneous medical advice).
Detecting and quantifying hallucination risk is thus a critical capability to enable safe applications of LLMs and improve generated outputs.

\begin{figure}
    \centering
    \includegraphics[width=\linewidth]{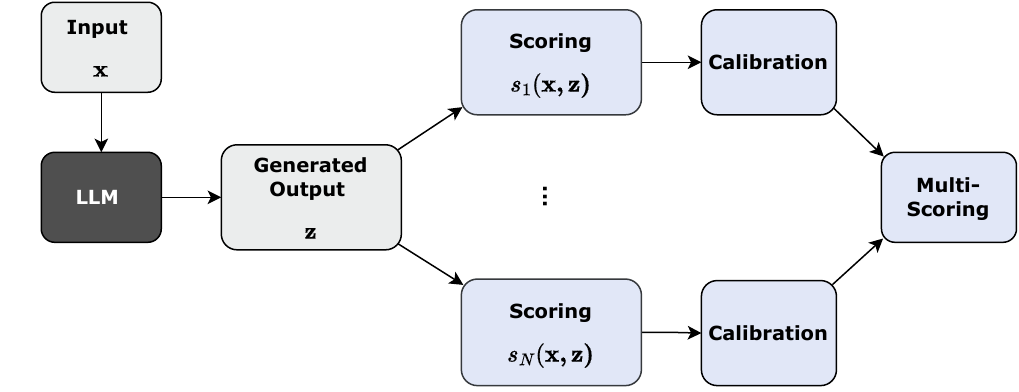}
    \caption{Schematic overview of our proposed hallucination detection approach.}
    \label{fig:enter-label}
\end{figure}

Prior work has proposed various approaches for detecting and mitigating hallucinations in LLM-generated outputs, including verifying faithfulness to inputs~\cite{maynez2020faithfulness}, assessing internal coherence~\cite{liu2021token}, consulting external knowledge sources~\cite{shuster2021retrieval}, and quantifying model uncertainty~\cite{xiao2021hallucination, zhang2023siren, ji2023survey, detommaso2024multicalibration}.
However, deploying these methods in production settings is far from trivial due to several challenges:
First, there is limited comparative evaluation illuminating how different detection methods perform. %
Second, existing approaches for detecting hallucinations differ greatly in their computational demands, and guidelines are lacking on cost-effectiveness trade-offs to inform method selection for real-world applications with constraints.
Third, hallucination detection in the real world often requires careful consideration of risks and false positive/negative trade-offs, requiring methods to provide well-calibrated probability scores.
Fourth, many applications of LLMs take the form of calls to black-box APIs, which sometimes requires methods to employ workarounds to assess the model's confidence in its generated output.

In this work, we provide a framework for detecting hallucinations in the outputs of any LLM in a model-agnostic manner.
Our approach relies on quantifying the probability that a generated answer is a hallucination.
After computing initial scores, we employ state-of-the-art calibration techniques to provide calibrated probabilities of the generation containing a hallucination, which can subsequently be used for decision making or other downstream tasks.
We evaluate a variety of scores proposed in the literature for hallucination detection on several metrics, across multiple datasets encompassing question answering, fact checking, and summarization tasks. We employ a range of different LLMs to ensure a comprehensive assessment of performance.

Critically, as we show that no single score performs best across all datasets,
we introduce
\emph{multi-scoring},
a simple way of combining multiple calibrated scores, which achieves superior performance compared to
any individual score alone.
Furthermore,
we
propose \emph{cost-effective multi-scoring}, which
finds the subset of best-performing scores for any fixed cost budget, and combines them in a multi-scoring fashion. Our empirical demonstrations reveal that cost-effective multi-scoring not only matches but often surpasses the performance of individual scores that incur significantly higher costs. Consequently, our proposed method achieves superior hallucination detection outcomes while maintaining a substantially lower cost footprint.

We summarize our contributions as follows: 1. We benchmark a variety of hallucination detection methods across the literature on several metrics, over different datasets and LLMs. 2. We introduce multi-scoring, a novel approach that aggregates multiple complementary scores and outperforms individual scores. 3. We further propose cost-effective multi-scoring, which optimally balances detection performance and computational constraints. %

\section{Detecting LLM Hallucinations}
\label{sec:formalization}
\subsection{Formalizing Hallucination Detection}
We study the problem of quantifying the probability that a generated output from a language model contains hallucinations.
More formally, let %
$\bm{x}$ represent an input token sequence to a language model $\mathcal{G}$, and let %
$\bm{z}$ represent a generated output text sequence from the model.
We define a binary random variable $y \in \{0,1\}$ that indicates whether $\bm{z}$ is a ``permissible'' output ($y=1$) or contains a ``hallucination'' ($y=0$).

Our goal is to develop a scoring function to model the probability that a given output text contains a hallucination conditioned on the input.\footnote{Note that while we are interested in whether the entire output contains any hallucination, one could easily adapt the methods at different granularities, such as the sentence or phrase level.}
This is critical, as in real-world scenarios, we need to set risk-aware thresholds, balancing false positive/negative rates to accommodate the production requirement.
Having access to scores allows us to flexibly set the threshold.
Conceptually, the key reason to model this probabilistically is that there is inherent uncertainty in determining whether a given text contains a hallucination or not.
Even human raters may disagree on the assessment, based on their own knowledge and definition.
Some key contributors to this epistemic uncertainty are: First, that no system has complete world knowledge to perfectly assess factual correctness.
Second, that there is ambiguity in whether something is a hallucination. %
Finally, any automatic scoring model may make occasional errors, so probabilistic scores reflect confidence.
Therefore, while conditional on $\bm{x}$ and $\bm{z}$ the true hallucination label $y$ is fixed, our estimate of $y$ remains uncertain. The probabilistic score thus reflects this epistemic uncertainty --- the degree of belief that $z$ contains a hallucination given the available knowledge:
$p(y = 0 \mid \bm{x}, \bm{z}; \bm{\psi})$, where $\bm{\psi}$ denotes parameters of the scoring function. We refer to this conditional probability function as the hallucination score, denoted $s_{\bm{\psi}}(\bm{x}, \bm{z})$.
The hallucination score can then be applied to downstream tasks, including making risk-aware binary decisions.

We discuss concrete instances of hallucination scores in the following subsection.
Generally, the form of $s_{\bm{\psi}}$ can vary between hallucination detection approaches.
In particular, $s_{\bm{\psi}}$ may depend on multiple candidate texts generated for the same input, as implemented by a number of hallucination detection methods proposed in the literature.
We denote $K$ candidate generated outputs as $\bm{Z} = [\bm{z}_1,...,\bm{z}_K]$.
Then $s_{\bm{\psi}}(\bm{x}, \bm{Z})$ could quantify inconsistencies within texts in $\bm{Z}$ using different metrics $\bm{\psi}$, as discussed below.

\subsection{Scoring Methods}
Many hallucination detection methods proposed in the literature make use of LLMs to ``judge'' the output of an LLM (either the same or a different one).
For clarity, we thus distinguish between \emph{generator} and \emph{detector} LLMs.
The \emph{generator} is defined as the model used to generate the original response. %
The \emph{detector} LLM is the model used to score a generated text for the presence of hallucinations.
In general, the \emph{generator} and \emph{detector} LLMs may coincide.
However, it may be more desirable to use different LLMs in some scenarios, e.g., when computational cost is a greater concern, where a smaller LLM may be used to judge outputs of a more expensive LLM, or when using hallucination scoring methods that require white-/grey-box access, while the generator is black-box.

Today, many interactions with LLMs take the form of API-calls.
Typically, only the output tokens are returned with no access to the logits of the predicted tokens,  treating the LLM effectively as a black box.
Sometimes, inference parameters may be accessible, allowing for setting different temperature (among others) values, thereby providing some (grey-box) access to the model.
Generally, hallucination detection methods vary in their required model access, ranging from black-box APIs to full white-box access to the model weights.
That is, while some methods only require token-level outputs, other methods may need access to the logits of the generated tokens, or some control over inference parameters like temperature.

In our experiments, we evaluate a comprehensive set of hallucination scoring methods.
Generally, we do not make any assumptions about the generator LLM being white-, grey- or black-box.
We divide methods into \emph{single-generation} methods, which require only one generated output, and \emph{multi-generation} methods which are based on multiple alternative generations.

\begin{table*}
    \centering
    \small
    \setlength{\tabcolsep}{4pt}
    \begin{tabularx}{\textwidth}{lccccX}
    \toprule
    \textbf{Score} & \textbf{\makecell{Logit\\Access}} & \textbf{\makecell{NLI\\model}} & \textbf{\makecell{\# LLM\\calls}} & \textbf{\makecell{\# NLI\\calls}} & \textbf{\# Generations} \\
    \midrule
    P(True) & Yes & No & 1 & 0 & 0 \\
    P(True) Verbalized & No & No & 1 & 0 & 1 \\
    P(InputContradict) & Yes & No & 1 & 0 & 0 \\
    P(SelfContradict) & Yes & No & 1 & 0 & 0 \\
    P(FactContradict) & Yes & No & 1 & 0 & 0 \\
    Inverse Perplexity & Yes & No & 1 & 0 & 0 \\
    NLI (DeBERTa) & No & Yes & 0 & 1 & 0 \\
    SelfCheckGPT-NLI & No & Yes & $K$ & $MK$ & $K$ \\
    HallucinationRail & Yes & No & $K$ & 0 & $K$ \\
    SimilarityDegree & No & Yes & $K$ & $K$ & $K$ \\
    \bottomrule
    \end{tabularx}
    \caption{Summarizing properties of different scoring methods including model access and inference time costs. $K$ denotes the number of multiple generations, $M$ denotes the number of sentences in the response. The column of \# Generations denotes the number of LLM calls required besides the LLM call that generates the original response. In this work we set $M = 1$ as we consider the responses as one sentence in our experiment.}
    \label{tab:score_properties}
\end{table*}

\subsection{Single-generation}
We first provide an overview over a set of hallucination detection methods that are based on scoring the hallucination from a single given generated output.

\paragraph{Inverse Perplexity} This method provides a prominent instance of using a model's logits over output tokens to score the confidence in the output. Computed as the inverse of the exponentiated average negative log-likelihood of the LLM's response~\cite{jelinek1977perplexity},
\begin{equation}
\label{eq:inv_perp}
    \text{Perplexity}^{-1}(W) = \exp\left(\frac{1}{N}\sum_{i=1}^N \log p(w_i|w_{i-1},...,w_{1})\right),
\end{equation}
inverse perplexity thus provides a sequence-length normalized expression of the model's confidence.
Here, the model may either be the generator LLM or a different detector LLM.
Using a different LLM from the generator as a plug-in estimator implies that the confidence estimate is detached from the generator.
If our generator provides access to the model's logits we can directly use them to score hallucinations.
Alternatively, we can use another LLM to generate the logits of output tokens.

\paragraph{P(True)} This method works by prompting an LLM whether an answer is correct or not, then making use of the logits of the next token~\cite{kadavath2022language}.
Given a prompt like the following:
\begin{verbatim}

Provide a "True" or "False" response on whether the
answer for the following question is correct. Give ONLY
a True or False answer, no other words or explanation.

The question is: {x}
The answer is: {z}
The answer is:

\end{verbatim}

We compute P(True) by first taking the softmax over the first generated output token's logits,
and then normalizing across the ``True'' and ``False'' tokens.\footnote{Note that alternatively, we can prompt the LLM to bind the True/False tokens to symbols like response options A or B~\cite{kadavath2022language}. However, this did not change results in our experiments. Further, if the detector LLM provides only black-box access, one can sample the corresponding tokens and use the empirical distribution to approximate, but thereby incurring a higher computational cost.}
That is, let $\mathbf{z}^{(1)} = (z^{(1)}_1, \ldots, z^{(1)}_{|V|})$ be the logit vector over the vocabulary $V$ for the first generated token when asking the LLM to evaluate whether the answer is correct. We then take the softmax over these logits,
$p(w \mid \mathbf{z}^{(1)}) = \frac{\exp(z^{(1)}_w)}{\sum_{w' \in V} \exp(z^{(1)}_{w'})}$.
Let $w_{\text{True}}, w_{\text{False}} \in V$ be the ``True'' and ``False'' token ids. Then we can compute P(True) as:
\begin{equation}
\label{eq:p_true}
P(\text{True} \mid \mathbf{z}^{(1)}) = \frac{p(w_{\text{True}} \mid \mathbf{z}^{(1)})}{p(w_{\text{True}} \mid \mathbf{z}^{(1)}) + p(w_{\text{False}} \mid \mathbf{z}^{(1)})}.
\end{equation}

In addition to checking whether the response is correct, we provide variations of P(True), where we check whether the output contradicts the input in \emph{P(InputContradict)}, the output contradicts itself in \emph{P(SelfContradict)}, or the output contradicts generally established facts in \emph{P(FactContradict)}, with the followings prompts: %
\begin{verbatim}
P(InputContradict): Provide a “True“ or ”False“ response
on whether the following texts are free of any direct
logical or factual contradictions between them. Give
ONLY a  True or False answer, no other words or explanation.

First text: {x}
Second text: {z}
The answer is:
\end{verbatim}

\begin{verbatim}
P(SelfContradict): Provide a “True“ or ”False“ response
on whether the  following text is free of internal
factual or logical contradictions. Give ONLY a True or
False answer, no other words or explanation.

The text is: {z}
The text is internally consistent:
\end{verbatim}

\begin{verbatim}
P(FactContradict): Provide a “True“ or ”False“ response
on whether the following text contains no contradictions
with generally established facts. Give ONLY a True or
False answer, no other words or explanation.

Text: {z}
The text is factually sound:
\end{verbatim}

\paragraph{NLI Text Classification} Natural language inference (NLI) models provide an alternative to assess the correctness of the model output. As hallucination detection requires checking for contradictions, we compute the score as $1-S_\text{contradict}$, where $S_\text{contradict}$ refers to the softmax probability of the output conflicting with the question. Specifically, we use a DeBERTa model fine-tuned on an MNLI task~\cite{he2020deberta} as the underlying NLI model, which we refer to as \emph{NLI (DeBERTa)}.

\paragraph{Verbalized Probabilities} Instead of analyzing a model's logits, scores are elicited by asking an LLM to provide a confidence \emph{verbatim}, i.e., by generating tokens that indicate the numerical confidence~\cite{lin2022teaching, tian2023just} with the following prompt:
\begin{verbatim}

Provide the probability between 0.0 and 1.0 that the
answer for the following question is correct. Give
ONLY the probability value between 0.0 and 1.0, no
other words or explanation.

The question is: {x}
The answer is: {z}
Probability the answer is correct:

\end{verbatim}
We note that moderated LLMs accessed through APIs may in practice decline to answer if the context does not provide sufficient information, rather than assigning a low confidence score.

\subsection{Multi-generation} Multi-generation methods are based on quantifying the consistency across multiple generated outputs from the generator LLM.
This follows the notion of white-/grey-box uncertainty quantification via logits that if the LLM is confident in its response, multiple generated responses will probably be alike and contain compatible facts.
Conversely, for fabricated information, sampled responses are more likely to differ and contradict one another. Crucially, this rests on the assumption that the model's confidence is calibrated, as we discuss and evaluate below.

\paragraph{SelfCheckGPT} This method exploits a DeBERTa NLI model to assess whether the answer $A_0$ is consistent with $K$ alternative generated answers $A_1, ..., A_K$~\cite{manakul2023selfcheckgpt}. Each sentence of $A_0$ is compared against the full set of answers $A_1, ..., A_K$. A consistency score per sentence is obtained via an NLI model, then the final score is obtained by averaging over the sentence-wise scores. %

\paragraph{Similarity Degree} This method is based on computing pairwise similarities between the multiple responses using NLI models, and then quantifying uncertainty based on the distribution of similarities~\cite{lin2023generating}. Here, we compute the pairwise similarities between responses via the \emph{contradict} class probability of an NLI model and construct a degree matrix, where each diagonal element corresponds to the total (sum) similarity of the corresponding response to all other responses. We use the degree of the candidate response $A_0$ as the confidence estimate, following favourable results in prior work~\cite{lin2023generating}, but note that other metrics have also been proposed in the original paper.

\paragraph{NeMO Guardrails: Hallucination Rail} %
Unlike SelfCheckGPT or Similarity Degree, the score here is computed via one LLM call, where we check for agreement between the concatenated $K$ additional generations and the candidate output and not averaged over sentences or computed on pairs of responses. Specifically, we use the softmax probability normalized over yes/no tokens using the following prompt:

\begin{verbatim}
You are given a task to identify if the hypothesis is
in agreement with the context below. You will only
use the contents of the context and not rely on
external knowledge. Answer with yes/no.

context: {K additional_sampled_responses}
hypothesis: {candidate_response}
agreement:
\end{verbatim}

\subsection{Calibration}
Initial hallucination scores may not be properly calibrated, which can lead to poor downstream decisions. Seminal work by Guo et al.~\cite{guo2017calibration} has demonstrated that neural models tend to be \emph{miscalibrated}, particularly in the form of models being overconfident.
Later work, focusing on language models, has confirmed this across a wide range of tasks, models and datasets~\cite{lin2023generating}.
Notably, there is also research which demonstrates that given particular prompts and in-distribution tasks, LLMs can be well-calibrated~\cite{kadavath2022language}. However, this has been shown to be brittle and dependent on context~\cite{kuhn2023semantic}.

Formally, our score outputs probabilities $p_{\bm{\psi}}(\hat{y}=0|\bm{x}, \bm{z})$ that $\bm{z}$ contains a hallucination, parameterized by $\bm{\psi}$.
The score is calibrated if, for any probability level $p \in [0,1]$, the average observed frequency of hallucinations matches the predicted probability:
\begin{equation}
\label{eq:calib}
    \mathbb{E}[y \mid p_{\bm{\psi}}(\hat{y}=0|\bm{x}, \bm{z}) = p] = p
\end{equation}

A naive approach for obtaining scores would be to compute the model's probabilities marginally, ignoring the context/question $\bm{x}$ and generated response $\bm{z}$ and directly estimating $p(y=1)$.
However, this does not allow for conditioning on individual inputs to obtain calibrated probabilities $p(y=1 \mid \bm{x}, \bm{z})$ for specific $\bm{x}$, and $\bm{z}$, which is, however, impossible to guarantee~\cite{roth2022uncertain}.
Common calibration methods include temperature scaling (Platt scaling) of logit outputs~\cite{guo2017calibration, platt1999probabilistic}, isotonic regression~\cite{zadrozny2002transforming} or histogram binning~\cite{zadrozny2001obtaining}, which operate marginally, and can thereby not account for different confidence levels for different inputs (e.g., with an LLM being more confident on certain domains than others).

An alternative is to partition the inputs into $G$ groups and compute calibration separately for each group $g\in G$. However, this assumes the groups are disjoint and does not handle inputs belonging to multiple groups, which is often necessary~\cite{roth2022uncertain}.
More advanced calibration methods, such as \emph{multicalibration}, which we use in this work, allow defining $G$ potentially overlapping groups ~\cite{roth2022uncertain}.
Prior work has shown the effectiveness of modern calibration techniques in scoring LLM confidence, albeit in a white-box setting~\cite{detommaso2024multicalibration}.
We describe our calibration approach in the experimental section.

\subsection{Multi-Scoring: Combining Scores}
Different scoring methods capture different aspects of hallucinations, e.g., incorrect, non-factual, non-grounded, irrelevant, inconsistent with other answers, etc. As a result, some scores may work better on some data or for some specific models, and worse on others. Therefore, we design a \emph{multi-scoring} method to combine the complementary information from individual scores into a single, strong predictor.

Denote each available score by $s_n$, for $n=1, \ldots ,N$. To obtain an aggregated score, we run a logistic regression using as features the concatenations of the logit of each score, i.e. $\left[\text{logit}(s_n(\bm{x}, \bm{z})), \ldots, \text{logit}(s_N(\bm{x}, \bm{z})) \right]$, and the labels of the calibration dataset as target variables.~\footnote{Note that we also conducted experiments with alternative models, such as XGBoost~\cite{chen2016xgboost} and Random Forest~\cite{breiman2001random}, but results did not show significant improvements, so we opted for logistic regression for simplicity.}

\subsection{Cost-Effective Multi-Scoring}
Scoring methods based on multiple generations incur the cost of additional generations as well as the cost of checking their consistency.
Similarly, multi-scoring can be a viable choice when there are only few LLM generations to check for hallucinations, but incurs considerable computational cost, especially when using multi-generation methods.
To avoid prohibitive computational costs in practice, we propose \emph{cost-effective multi-scoring}, where we set a fixed computational budget and compute the best performing combination of scores that stay within the specified budget.

Given an input text $\bm{x}$ and generated text $\bm{z}$, we have $N$ scoring functions $s_1(\bm{x}, \bm{z}), ..., s_N(\bm{x}, \bm{z})$ with associated costs $c_1, ..., c_N$ (e.g., number of generations required). We are given a total computational budget $B$. Our goal is to find the optimal subset of scores $S^* \subseteq \{1,...,N\}$ that maximizes detection performance while staying within budget $B$:

\begin{equation}
\label{eq:cost_aware_opt}
    S^{\ast} = \argmin_{S: S \subseteq \{1, \ldots, N\}} \mathcal{L}(f({s_i(\bm{x}, \bm{z}))}_{i \in S}) \quad \text{s.t.} \quad \sum_{i \in S} c_i \leq B
\end{equation}

where $\mathcal{L}$ measures loss on a validation set. This is a constrained optimization problem over subsets of scoring functions. When $B=\min_i c_i$, it reduces to selecting the single best score at the lowest cost. When $B=\sum_i c_i$, it recovers full multi-scoring. In between, the optimal subset $S^*$ provides the best trade-off between performance and cost.
We note that in general, this problem is computationally challenging, given the exponential runtime complexity.
However, given that there are only generally a small set of potential scores ($N \sim 10$) and the logistic regression is very fast to fit, computing all combinations takes only $1.8$ seconds on a single Intel Xeon processor (3.1 GHz), iterating over all candidate solutions sequentially.
When this approach is not feasible, some exemplary alternatives include classic greedy forward-selection methods or regularising the model via an L1 penalty while scaling the regularisation term to accommodate score cost.
Overall, cost-effective multi-scoring allows for flexibly balancing multiple scores under computational constraints.

\paragraph{Quantifying Cost}
While the cost $c_i$ of scoring function $s_i$ is presented abstractly above, accurately quantifying computational cost can be challenging in practice.
The actual runtime per method is a reasonable proxy, however the runtime depends on model architecture, hardware acceleration, batching, etc.
One approach is to directly benchmark each scoring function's average runtime empirically on the target hardware.
However, this overlooks nuances like caching effects and ignores runtime variability due to implementational differences.
To simplify our analysis, and as LLM calls are generally much more expensive than calling smaller NLI models based on parameter size, we leverage the number of LLM calls required per method as a proxy. See Table~\ref{tab:score_properties} for an overview.
If more precision is desired, we suggest empirically running different scores and computing their computational cost in the actual application, as the cost will depend on the precise setup and a range of factors.

\section{Experiments}
\label{sec:experiments}
\subsection{Experimental Setup}
\subsubsection{Datasets}

\paragraph{TriviaQA} TriviaQA~\cite{joshi2017triviaqa} is a commonly used factual open-respon\-se QA dataset. Originally set up as a reading comprehension task, it is today often used without context as a closed-book (free-recall) task~\cite{wei2021finetuned}.
We use the validation fold, containing $17944$ question-answer pairs. We use \texttt{mistralai/Mistral-7B-Instruct\-v0.2}~\cite{jiang2023mistral} to generate candidate answers. To decide whether a given response should be labelled as positive or negative, we check whether the correct answer is contained in the generated answer after removing formatting, following the original evaluation script~\cite{joshi2017triviaqa}.

\paragraph{FEVER} The closed-response Fact Extraction and VERification dataset~\cite{thorne2018fever} provides a comprehensive benchmark of factual hallucination detection. We take the test fold, containing $14027$ labelled examples of source documents, claims and whether they are  \texttt{supported}, \texttt{refuted} or contain \texttt{not enough info}. For the purpose of hallucination detection, we look at all claims that are either \texttt{supported} or \texttt{refuted}.%

\paragraph{HaluEval} We include the hallucination detection dataset HaluEval~\cite{li2023halueval} and use the summarization task, which contains $10000$ labeled examples of source documents, summaries and labels for whether the summary contains hallucinations.

\paragraph{BIG-bench} It provides an evaluation of LLMs on a diverse set of tasks~\cite{srivastava2022beyond}. We select 11 tasks suitable for hallucination detection, leading a total of $8664$ labelled examples from the validation fold.

\subsubsection{Metrics}
Practical applications of hallucination detection require calibrated scores, but often also binary decisions over whether a given output is hallucinated or not.
To this end we report the Brier score~\cite{brier1950verification}, binary decision metrics F1 score and accuracy.

\subsubsection{LLMs}
We conduct experiments with \texttt{Mistral-7B-Instruct\-v0.2}~\cite{jiang2023mistral}, \texttt{Mixtral-8x7B-Instruct-v0.1}~\cite{jiang2024mixtral}, \texttt{falcon\-7b-instruct}~\cite{falcon40b} and \texttt{OpenLLaMA-13b}~\cite{together2023redpajama, touvron2023llama, openlm2023openllama}.
All models are used with default configurations via HuggingFace Transformers~\cite{wolf2019huggingface}.

\subsubsection{Calibration}
The calibration step is performed via the following multicalibration approach, using the Fortuna library~\cite{detommaso2023fortuna}:
To obtain groups, we compute embeddings of the input text $\bm{x}$ and the generated response $\bm{z}$, such that our embedding is $\bm{e} \coloneqq \left[ \texttt{embed}(\bm{x}), \texttt{embed}(\bm{z}) \right]$. We obtain embeddings from Universal AnglE Embedding~\cite{li2023angle}, which is the SOTA in the MTEB benchmark~\cite{muennighoff2022mteb} at the time of writing.
We subsequently reduce the dimension of $\bm{x}$ via UMAP~\cite{mcinnes2018umap-software} and perform soft-clustering via Gaussian Mixture Models, as a simple off-the-shelf algorithm (fitting the number of cluster components via BIC~\cite{schwarz1978estimating}).
The calibration error is measured for each group $g\in G$ separately.
Calibration then involves iteratively patching the group with the largest error until the calibration error drops below a threshold for all groups.\footnote{Note that alternative approaches are possible~\cite{roth2022uncertain}.}
This provably converges to a calibrated model with theoretical guarantees~\cite{roth2022uncertain}.
We fit the calibrator to a random \emph{calibration} fold of 80\% of the data and report only held-out test results.
For all binary predictions, we set the threshold to the $50$th percentile to not impose preferences over false true/negative rates, but note that in practice any threshold could be applied.

\subsection{Individual Scoring Methods}
Table~\ref{tab:res} presents the results of different hallucination detection methods on all datasets. Multi-generation methods (SelfCheckGPT-NLI, HallucinationRail and SimilarityDegree) are employed only for TriviaQA, since the latter provides the true answer to each question, which can be compared to the alternative generated answers to assess their agreement. In contrast, for HaluEval, BIG-Bench and FEVER, the true answer is not provided. Instead, we compare the given candidate answer from the dataset against the provided binary label of correctness.
If we exclude Multi-Scoring methods, multi-generation methods such as SelfCheckGPT-NLI and SimilarityDegree (with 10 generations) achieve the best performance on TriviaQA, P(True) is the best on HaluEval and BIG-Bench, and P(InputContradict) performs best on FEVER.
Thus, we find that there is no single best scoring method across all datasets.
This is likely because different scoring methods capture different aspects of hallucination, supporting the notion that hallucination is a multi-faceted concept.
Multi-generation methods measure hallucination based on the consistency of the generator LLM's responses and on the ability of the comparison method to identify whether multiple alternative responses are in agreement.
We find that SelfCheckGPT-NLI performs best in our experiments in comparison to SimilarityScore and HallucinationRail.
While all of these methods follow similar approaches, there are subtle differences in how they assess the consistency between different responses.
More generally, as we discuss below, multi-generation methods are appropriate only if one can assume that there is exactly one correct response, but can fail otherwise.

Other methods are based on different notions of hallucination, such as NLI (DeBERTa) measuring the entailment of the response given the input.
Interestingly, variants of P(True) can be appropriate for detecting different kinds of hallucinations.
P(True) explicitly asks the evaluator LLM to check whether the response is correct.
In some contexts, the evaluator LLM may have the ability to directly evaluate this.
However, in other situations, as we see with the FEVER dataset, other methods such as P(InputContradict) can be more appropriate, if we are trying to directly target a specific form of hallucination, which may not be subsumed under a general ``correct or not?'' prompt.
Other applied scenarios may target even more different (or more specific) kinds of hallucinations, though the variants we include in this work are designed to cover the space in a reasonable manner.
Similarly, while the datasets included in these experiments were selected to cover a wide range of different kinds of hallucinations, real-world applications may show even new kinds of hallucinations, for which no public datasets are available.
Overall, these findings highlight the need for our multi-scoring method which can absorb the strength of each individual method and can be easily applied to a concrete hallucination detection setting while only requiring a relatively small amount of labeled data.

Overall, neither the inverse perplexity score nor the NLI (DeBERTa) scores emerge as the best scores for any of the datasets we consider. While they may add information that can be exploited in (cost-effective) multi-scoring, as reported below, individually they perform worse than some of the other scores.

\begin{table*}
\centering
\caption{Hallucination detection results on all datasets of calibrated scoring methods. Methods that are not applied to a given dataset are marked as ---. All scores were computed with \texttt{Mistral-7B-Instruct-v0.2}. TriviaQA responses were generated with \texttt{Mistral-7B-Instruct-v0.2}. Excluding multi-scoring, the best performing scores are underlined. Including multi-scoring, the best performing scores are displayed in boldface.}
\label{tab:res}
\resizebox{\textwidth}{!}{
\begin{tabular}{@{}lllllllllllll@{}}
\toprule[1pt]
                        & \multicolumn{3}{l}{\textbf{TriviaQA}} & \multicolumn{3}{l}{\textbf{HaluEval}} & \multicolumn{3}{l}{\textbf{BIG-Bench}} & \multicolumn{3}{l}{\textbf{FEVER}} \\ \midrule
\textbf{Scoring Method} & Brier $\downarrow$    & F1 $\uparrow$     & Acc $\uparrow$    & Brier $\downarrow$  & F1 $\uparrow$    & Acc $\uparrow$   & Brier $\downarrow$    & F1 $\uparrow$     & Acc $\uparrow$    & Brier $\downarrow$   & F1 $\uparrow$ & Acc $\uparrow$   \\ \midrule
\midrule
P(True)                 & 0.1819 & 0.8263 & 0.7490 & \underline{0.1980} & \underline{0.7595} & \underline{0.6935} & \underline{0.2066} & \underline{\textbf{0.6470}} & \underline{0.6417} & 0.0729 & 0.9181 & 0.9180  \\
P(True) Verbalized      & 0.1789  & 0.8143 & 0.7350 & 0.2293  & 0.7040 & 0.6035 & 0.2149  & 0.5252 & 0.6255 & 0.0683  & 0.9197 & 0.9187  \\
P(InputContradict)      & 0.1799  & 0.8150 & 0.7398 & 0.2012  & 0.7552 & 0.6735 & 0.2375  & 0.0000 & 0.6042 & \underline{0.0634}  & \underline{0.9328} & \underline{0.9309}  \\
P(SelfContradict)       & 0.2125  & 0.8157 & 0.6888 & 0.2449 & 0.6639 & 0.5620 & 0.2423 & 0.1929 & 0.5413 & 0.1564  & 0.8216 & 0.8001  \\
P(FactContradict)       & 0.2107  & 0.8157 & 0.6888 & 0.2393  & 0.6409 & 0.6000 & 0.2355  & 0.2733 & 0.5949 & 0.1430 & 0.8207 & 0.8190  \\
\midrule
Inverse Perplexity      & 0.2033 & 0.8157 & 0.6888 & 0.2490 & 0.5151 & 0.5060 & 0.2289  & 0.2491 & 0.6244 & 0.2353 & 0.5653 & 0.5916 \\
\midrule
NLI (DeBERTa)           & 0.1924  & 0.8539 & 0.7451 & 0.2417  & 0.6602 & 0.5640 & 0.2322 & 0.3262 & 0.6186 & 0.1673 & 0.7512 & 0.7530  \\
\midrule
SelfCheckGPT-NLI (10)     & \underline{0.1434} & \underline{0.8614} & \underline{0.8011} & ---   &  ---    &  ---    & ---   &  ---    &  ---    &  ---    &  ---     &  ---     \\
HallucinationRail (10)    & 0.1640 & 0.8539 & 0.7721 & ---   &  ---    &  ---    & ---   &  ---    &  ---    &  ---    &  ---     &  ---       \\
SimilarityDegree (10)       &0.1443  & 0.8585 & 0.7927 & ---   &  ---    &  ---    & ---   &  ---    &  ---    &  ---    &  ---     &  ---      \\
\midrule
\midrule
Multi-Score             & \textbf{0.1105}  & \textbf{0.9106} & \textbf{0.8593} & \textbf{0.1911} & \textbf{0.7668} & \textbf{0.7075}  & \textbf{0.2045}  & 0.5966 & \textbf{0.6590} & \textbf{0.0544}  & \textbf{0.9371} & \textbf{0.9351}        \\
\midrule
Cost-Effective (C = 1)  & 0.1819  & 0.8263 & 0.7490 & 0.1980 & 0.7595 & 0.6935  & 0.2066  & \textbf{0.6470} & 0.6417 & 0.0634 & 0.9328 & 0.9309  \\
Cost-Effective (C = 2)  & 0.1772 & 0.8308 & 0.7520 & \textbf{0.1911} & \textbf{0.7668} & \textbf{0.7075}  & \textbf{0.2045} & 0.5966 & \textbf{0.6590} & 0.0636 & 0.0005 & 0.9328  \\
Cost-Effective (C = 3)  & 0.1727 & 0.8324 & 0.7537 & \textbf{0.1911} & \textbf{0.7668} & \textbf{0.7075}  & \textbf{0.2045} & 0.5966 & \textbf{0.6590} & \textbf{0.0544}  & \textbf{0.9371} & \textbf{0.9351} \\
Cost-Effective (C = 4)  & 0.1718  & 0.8277 & 0.7481 & \textbf{0.1911}  & \textbf{0.7668} & \textbf{0.7075}  & \textbf{0.2045} & 0.5966 & \textbf{0.6590} & \textbf{0.0544}  & \textbf{0.9371} & \textbf{0.9351} \\
Cost-Effective (C = 5)  & 0.1485  & 0.8619 & 0.8011 & \textbf{0.1911}  & \textbf{0.7668} & \textbf{0.7075}  & \textbf{0.2045}  & 0.5966 & \textbf{0.6590} & \textbf{0.0544}  & \textbf{0.9371} & \textbf{0.9351} \\
\bottomrule[1pt]
\end{tabular}%
}
\end{table*}

\subsection{Multi-Scoring}
To evaluate the proposed multi-scoring method, we conduct experiments on all datasets combining the scores via logistic regression.
As presented in Table~\ref{tab:res}, the multi-scoring ensemble achieves an F1 score of $0.9106$ on TriviaQA, outperforming the best individual F1 score of $0.8614$ (achieved by SelfCheckGPT-NLI).
Similarly, Brier and Accuracy also show the highest performance for multi-scoring.
For HaluEval, we also find that multi-scoring outperforms the best individual score (P(True)) in all metrics.
BIG-Bench shows more nuance, as multi-scoring outperforms the best individual score (P(True)) on Brier and Accuracy, but not on F1.
This reflects the fact that the metrics capture different properties, and practical considerations may require a decision as to which metric should be prioritized in a given application.
For FEVER, multi-score again outperforms the best individual score (P(InputContradict)) in all metrics.

Thus we see that combining scores performs better than the top performing individual scores, showing that combinations of different scoring signals complement each other and can boost performance.
This demonstrates that combining complementary signals enables more robust hallucination detection, while balancing the strength of each scoring method.
Crucially, while some settings may benefit from combinations of different signals (such as when trying to detect different kinds of hallucinations in a given generated output), in other settings the data-driven selection of an informative signal may be sufficient (such as when trying to detect a more narrowly defined notion of hallucination).
These scenarios are covered by our use of multi-scoring, which allows for arriving at optimally combined hallucination scores (or binary decisions) for a given application.
Meanwhile, as computational cost can be a concern with scaling LLM applications, it may not be desirable to always use a full ensemble of scores, and we would rather compute the most performant score at a fixed computational cost.

\subsection{Cost-effective Multi-Scoring}
Shown in Table~\ref{tab:res}, cost-effective multi-scoring methods are also amongst the top performers, but at considerably lower cost compared to multi-score. We see that at cost $C=1$, measured as the number of LLM calls (but see our discussion about measuring cost above), we recover the best individual scores, which varies across each dataset. As we increase the budget, the cost-effective multi-scores converge to the performance of the multi-score itself. Notice that in several instances, cost-effective multi-score with $C=2$ already performs as good as multi-score itself.
For TriviaQA, we see that performance increases on all metrics as we increase the computational budget, with cost-effective multi-score at $C=5$ outperforming SelfCheckGPT-NLI as the most performant individual scores at half the computational cost in all metrics but Brier.
HaluEval shows no improvement beyond a budget of $C=2$, which is likely due to the kind of hallucination to detect being more narrowly defined and well-capture by a small number of signals, thereby not benefiting from additional scores.
For BIG-Bench, as discussed for multi-score, we recover the best individual score at $C=1$, and match multi-score at higher budgets.
FEVER results indicate that, again, we recover the best individual score at budget $C=1$, and can already recover the full multi-score at a budget of $C=3$.

\begin{figure}[]
\centering
\includegraphics[width=0.5\linewidth]{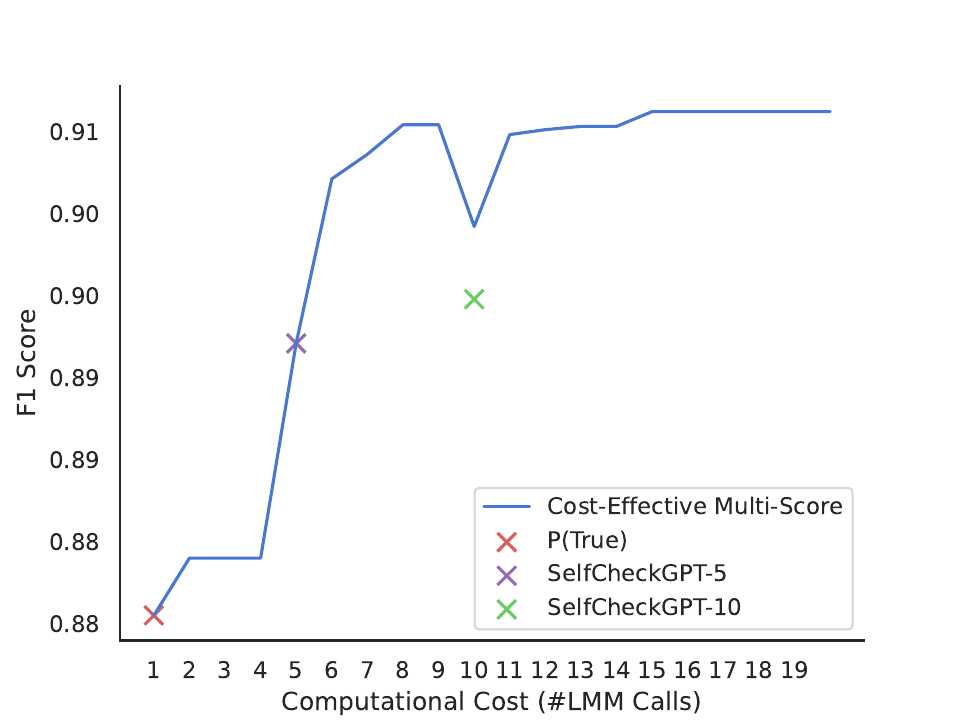}
\caption{Hallucination detection F1 versus computational budget $B$ for cost-effective multi-scoring.}
\label{fig:cost_aware_budget}
\end{figure}

We now take a closer look at cost-effect multi-scoring when including multi-generation scoring methods.
The overall cost budget is varied over the entire range. For each budget, we solve the constrained optimization in Eq.~\ref{eq:cost_aware_opt} to find the optimal subset $S^*$ of scores.
We compare the hallucination detection F1 score achieved by the cost-effective ensemble versus individual scoring functions and the full ensemble with all scores. Figure~\ref{fig:cost_aware_budget} shows the results for TriviaQA for responses generated via \texttt{Mixtral-8x7B-v0.1}. With a minimal budget of $B=1$, cost-effective selection recovers the best single method, as expected. As the budget increases, it selectively adds more expensive functions, gradually improving F1, though the gains are marginal at higher budgets per unit cost. At the maximum budget, it recovers the unconstrained full ensemble performance. In between, cost-effective multi-scoring incorporates both less and more and expensive methods to maximize detection within the computational constraints.

These results demonstrate that the proposed cost-effective multi-scoring approach can intelligently balance the trade-off between computational expense and hallucination detection effectiveness. It outperforms individual scoring functions and makes selective use of more costly scores to maximize detection performance under a fixed computational budget.

\begin{figure}[ht!]
    \centering
    \includegraphics[width=0.5\linewidth]{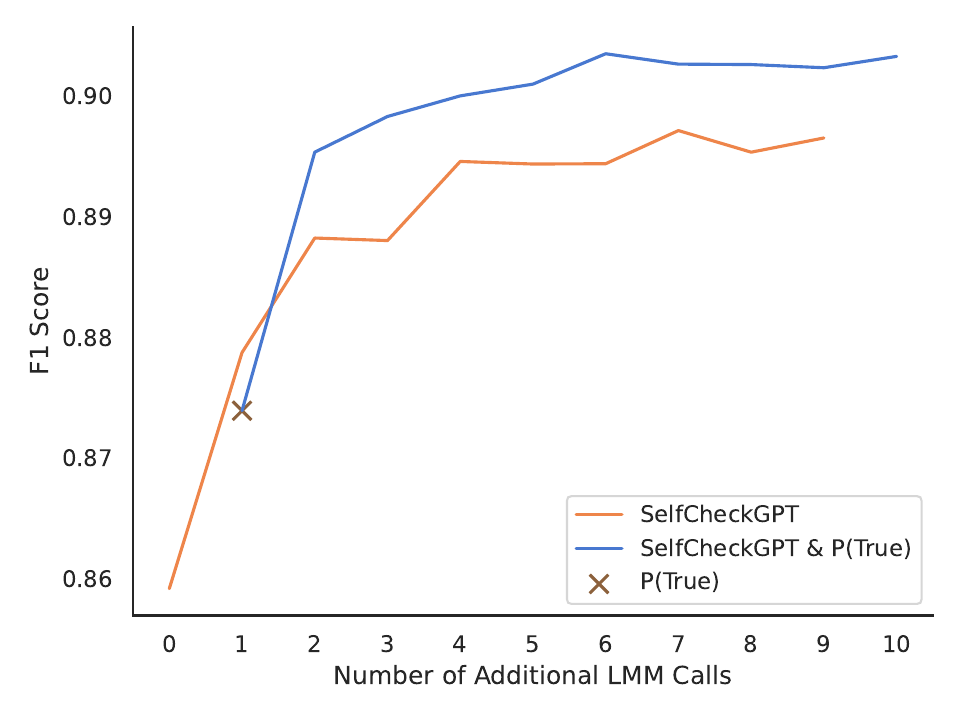}
    \caption{Relationship between number of generations used for SelfCheckGPT and performance of cost-effective multi-score vs SelfCheckGPT alone on TriviaQA.}
    \label{fig:cost_aware}
\end{figure}

A key consideration in practice would be whether to include potentially more costly multi-generation methods, such as SelfCheckG\-PT-NLI.
In particular as implied by the results presented in Table~\ref{tab:res}, we may be interested in reducing the number of required multiple generations while matching performance at lower computational costs.
To this end, we compare SelfCheckGPT (as the single best performing method for TriviaQA) alone with SelfCheckGPT combined with P(True) at different numbers of additional responses generated via \texttt{Mixtral-8x7B-v0.1}.
Here, we count the evaluation of P(True) as one additional LLM call, just as generating one additional output from the generator LLM.
As presented in Figure~\ref{fig:cost_aware}, at one additional generation, SelfCheckGPT (in the degenerate case of only one generated output) combined with P(True) thus recovers P(True).
As the number of additional generations increases, the combination quickly outperforms SelfCheck alone.
In particular, we observe that cost-effective multi-scoring with 3 LLM calls is already as good as SelfCheckGPT with 9 LLM calls in our experiment.
This highlights how we can save costs by combining multi-generation methods with other scores while requiring fewer additional generated responses from the generator LLM than if we wanted to achieve the same performance with multi-generation methods alone.

\subsection{Exploration of Relationships between Scores}

\begin{figure}
    \centering
    \includegraphics[width=0.5\linewidth]{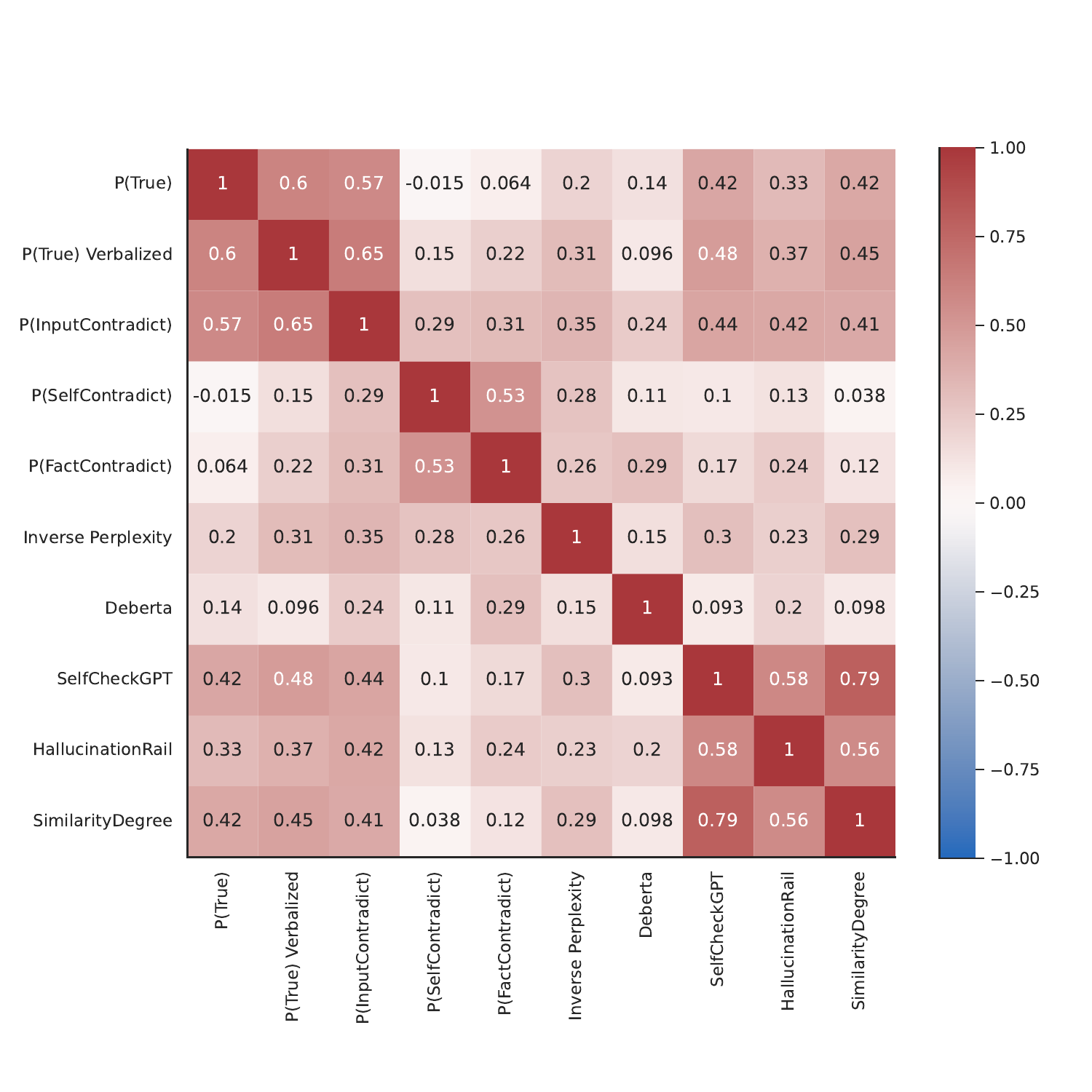}
    \caption{Heatmap of Spearman rank correlations between scores on TriviaQA.}
    \label{fig:score_corrs}
\end{figure}

As discussed above, different scoring methods can target different kinds of hallucations.
To explore this empirically, we here explore their relationships via Spearman rank correlations.
As presented in Figure~\ref{fig:score_corrs}, calibrated scores across all different individual scores are positively correlated.
Meanwhile, the magnitude of their correlations is smaller than would perhaps be  expected if one were to consider hallucinations as a uniform phenomenon.
At the same time, we can see certain ``clusters'' of more strongly inter-correlated scores emerge.
For example, multi-generation methods including SelfCheckGPT, HallucinationRail and SimilarityDegree, which check the consistency across multiple generations, show comparably high correlations.
P(True) and P(True) Verbalised and P(InputContradict) emerge as a similarly correlated cluster of stronger correlations.
Overall, this supports the idea that different scores can capture distinct information, and the need to empirically select appropriate (combinations of) scores for a given application, as we propose with cost-effective multi-scoring.

\subsection{Experiments across Different LLMs}
Most of the scoring methods considered in this work rely on detector LLMs to compute scores for hallucination detection.
In Table \ref{tab:res_mistral_7b} of Appendix \ref{apdx:add}, thus present additional data for different LLMs, namely \texttt{Mixtral-8x7B-Instruct-v0.1}, \texttt{falcon-7b-in\-struct} and \texttt{OpenL\-LaMA-13b}.
Overall, we see that hallucination detection performance is correlated with performance on general LLM benchmarks\footnote{E.g., \url{https://huggingface.co/spaces/HuggingFaceH4/open_llm_leaderboard}}.
Thus, more overall capable LLMs are likely to perform better at hallucination detection than less capable ones.
However, concrete applications may need to take the inference cost of LLMs into account, where one may not always be able to use the most expensive model.
We note that cost-effective multi-scoring could here also make use of scores computed via LLMs with different computational costs to find cost-effective combinations.

\subsection{The Importance of Calibration}
To analyze the impact of calibration, we conduct an ablation study by evaluating model performance with and without calibrating the individual scores.
Table \ref{tab:res_nocal} in Appendix \ref{apdx:add} indicates that model performance clearly drops in most of the datasets and metrics without calibration. This demonstrates the benefit of calibration for more accurate and risk-aware downstream applications.

\subsection{Exploring Multi-Generation Assumptions}

\begin{figure}[ht!]
    \centering %
    \begin{tcolorbox}[colback=gray!10, colframe=gray!20, boxrule=0.5pt, sharp corners]
        \textbf{User:} Generate a recipe for Lunch.\\
        \\
        \textbf{LLM Response 1:} Here is a recipe for a tomato and basil soup. First, bring 1L of vegetable broth to a boil...\\
        \\
        \textbf{LLM Response 2:} Here is a recipe for a quinoa and avocado salad. First, cook 150g of quinoa according to the package instructions and let it cool...
    \end{tcolorbox}
    \caption{Example of multi-generation failure-case in NLP systems, illustrating conflicting responses.}
    \label{fig:nlp_example}
\end{figure}

Our results indicate that methods based on the uncertainty among multiple generations can provide strong signals when such multiple generations are available.
However, such methods can also suffer from problems in practice.
Methods based on multiple generations make use of the uncertainty of the \emph{generator}, that is the distribution over generations $\bm{z}$ given the input $\bm{x}$, for model $\mathcal{G}$, i.e., $p_{\mathcal{G}}(\bm{z} \mid \bm{x})$ and the consistency between actual sampled generations $\bm{z}$.
We are interested in classifying whether the candidate output contains a hallucination or not,
that is estimating and making use of $p(y \mid \bm{x}, \bm{z})$.
The generating model's uncertainty $p_{\mathcal{G}}(\bm{z} \mid \bm{x})$ can be a useful proxy for $p(y \mid \bm{x}, \bm{z})$, but only under particular conditions.

On a technical level, sampling multiple answers requires access to the generator's temperature parameter, as temperatures of zero collapse the generations to a single response, as is mentioned, e.g., in~\cite{manakul2023selfcheckgpt}.
Therefore, these methods are technically grey-box models, as they require \emph{some} access to the model's inference parameters.

More conceptually, the generator's uncertainty (or confidence~\cite{lin2023generating}) over $p_{\mathcal{G}}(\bm{z} \mid \bm{x})$ reflects uncertainty over generated tokens.
All recently proposed multi-scoring methods we are aware of are based on the idea of scoring the consistency across multiple generations using model-based metrics.
However, self-consistency across multiple generations is neither a necessary nor sufficient criterion for a hallucination to be present.
Sufficiency is not given as a model may consistently provide an incorrect response.
Self-consistency across multiple generations is also not necessary, as many tasks allow for multiple hallucination-free answers that are contradictory to each other.

As a relatively harmless example, in tasking a model to generate cooking recipes for lunch, the model may generate, among other things, a recipe for a salad and a recipe for a soup. Clearly, the steps in preparing these dishes contain contradictory information, while otherwise being free of hallucinations in themselves.
Thus, if there is more than one correct response, the self-consistency assessment in multi-generation methods may falsely score an output as likely to be hallucinated.
As there are numerous situations where there exist multiple correct responses, assuming that there is only one could lead to worse LLM responses, also via a loss of diversity.

Practically, methods based on multiple generations can be costly due to added computational overhead, as we have seen above.
Finally, these methods are based on the assumption that LLMs are calibrated.
As such, multi-generation methods do not allow for detecting cases where the generator LLM is confident yet wrong.

\section{Conclusion}
\label{sec:conclusion}
In this work, we compared a comprehensive set of scoring methods to provide calibrated probability scores for the presence of hallucinations in generated LLM outputs.
Overall, we have observed that no single hallucination detection score performs best across all datasets.
Our experiments showed that combinations of scores, as suggested with multi-scoring, can effectively combine complementary signals to yield higher hallucination detection performance than any individual score.
Further, we demonstrate that cost-effective multi-scoring can find the highest performing scores at a given computational budget.

More generally, our findings support the notion that hallucinations can be rather multi-faceted than present a uniform phenomenon.
Thus, detecting hallucinations may require different methods.
In concrete settings, one may be interested in even more fine-grained detection of particular types of hallucinations, such as in specific domains like code generation~\cite{liu2024exploring}.

The approach outlined in this work requires only a small amount of labeled data to calibrate hallucination scores and combine scores via (cost-effective) multi-scoring.
However, it is important to acknowledge the limitations of the current work. Future research could explore the effectiveness of this approach on more diverse datasets, investigate alternative scoring methods, or extend the methodology to multi-modal models.
While improvements in LLM performance can be expected to also lower their rate of occurrence, hallucinations are unlikely to go away completely~\cite{kalai2023calibrated}.
Thus, even as models improve further, detecting hallucinations is likely to remain relevant for applying LLMs in a reliable and trustworthy manner in future.

In summary, our work presents a promising approach for detecting hallucinations in LLM outputs by combining multiple scoring methods in a cost-effective way, offering a pathway towards more reliable and trustworthy language models that can be applied in real-world settings with more confidence.

\bibliography{bibliography}

\begin{thebibliography}{10}

\bibitem{zhao2023survey}
Wayne~Xin Zhao, Kun Zhou, Junyi Li, Tianyi Tang, Xiaolei Wang, Yupeng Hou, Yingqian Min, Beichen Zhang, Junjie Zhang, Zican Dong, et~al.
\newblock A survey of large language models.
\newblock {\em arXiv preprint arXiv:2303.18223}, 2023.

\bibitem{ji2023survey}
Ziwei Ji, Nayeon Lee, Rita Frieske, Tiezheng Yu, Dan Su, Yan Xu, Etsuko Ishii, Ye~Jin Bang, Andrea Madotto, and Pascale Fung.
\newblock Survey of hallucination in natural language generation.
\newblock {\em ACM Computing Surveys}, 55(12):1--38, 2023.

\bibitem{zhang2023siren}
Yue Zhang, Yafu Li, Leyang Cui, Deng Cai, Lemao Liu, Tingchen Fu, Xinting Huang, Enbo Zhao, Yu~Zhang, Yulong Chen, et~al.
\newblock Siren's song in the ai ocean: A survey on hallucination in large language models.
\newblock {\em arXiv preprint arXiv:2309.01219}, 2023.

\bibitem{maynez2020faithfulness}
Joshua Maynez, Shashi Narayan, Bernd Bohnet, and Ryan McDonald.
\newblock On faithfulness and factuality in abstractive summarization.
\newblock {\em arXiv preprint arXiv:2005.00661}, 2020.

\bibitem{liu2021token}
Tianyu Liu, Yizhe Zhang, Chris Brockett, Yi~Mao, Zhifang Sui, Weizhu Chen, and Bill Dolan.
\newblock A token-level reference-free hallucination detection benchmark for free-form text generation.
\newblock {\em arXiv preprint arXiv:2104.08704}, 2021.

\bibitem{shuster2021retrieval}
Kurt Shuster, Spencer Poff, Moya Chen, Douwe Kiela, and Jason Weston.
\newblock Retrieval augmentation reduces hallucination in conversation.
\newblock {\em arXiv preprint arXiv:2104.07567}, 2021.

\bibitem{xiao2021hallucination}
Yijun Xiao and William~Yang Wang.
\newblock On hallucination and predictive uncertainty in conditional language generation.
\newblock {\em arXiv preprint arXiv:2103.15025}, 2021.

\bibitem{detommaso2024multicalibration}
Gianluca Detommaso, Martin Bertran, Riccardo Fogliato, and Aaron Roth.
\newblock Multicalibration for confidence scoring in llms.
\newblock {\em arXiv preprint arXiv:2404.04689}, 2024.

\bibitem{jelinek1977perplexity}
Fred Jelinek, Robert~L Mercer, Lalit~R Bahl, and James~K Baker.
\newblock Perplexity—a measure of the difficulty of speech recognition tasks.
\newblock {\em The Journal of the Acoustical Society of America}, 62(S1):S63--S63, 1977.

\bibitem{kadavath2022language}
Saurav Kadavath, Tom Conerly, Amanda Askell, Tom Henighan, Dawn Drain, Ethan Perez, Nicholas Schiefer, Zac Hatfield-Dodds, Nova DasSarma, Eli Tran-Johnson, et~al.
\newblock Language models (mostly) know what they know.
\newblock {\em arXiv preprint arXiv:2207.05221}, 2022.

\bibitem{he2020deberta}
Pengcheng He, Xiaodong Liu, Jianfeng Gao, and Weizhu Chen.
\newblock Deberta: Decoding-enhanced bert with disentangled attention.
\newblock {\em arXiv preprint arXiv:2006.03654}, 2020.

\bibitem{lin2022teaching}
Stephanie Lin, Jacob Hilton, and Owain Evans.
\newblock Teaching models to express their uncertainty in words.
\newblock {\em arXiv preprint arXiv:2205.14334}, 2022.

\bibitem{tian2023just}
Katherine Tian, Eric Mitchell, Allan Zhou, Archit Sharma, Rafael Rafailov, Huaxiu Yao, Chelsea Finn, and Christopher~D Manning.
\newblock Just ask for calibration: Strategies for eliciting calibrated confidence scores from language models fine-tuned with human feedback.
\newblock {\em arXiv preprint arXiv:2305.14975}, 2023.

\bibitem{manakul2023selfcheckgpt}
Potsawee Manakul, Adian Liusie, and Mark~JF Gales.
\newblock Selfcheckgpt: Zero-resource black-box hallucination detection for generative large language models.
\newblock {\em arXiv preprint arXiv:2303.08896}, 2023.

\bibitem{lin2023generating}
Zhen Lin, Shubhendu Trivedi, and Jimeng Sun.
\newblock Generating with confidence: Uncertainty quantification for black-box large language models.
\newblock {\em arXiv preprint arXiv:2305.19187}, 2023.

\bibitem{guo2017calibration}
Chuan Guo, Geoff Pleiss, Yu~Sun, and Kilian~Q Weinberger.
\newblock On calibration of modern neural networks.
\newblock In {\em International conference on machine learning}, pages 1321--1330. PMLR, 2017.

\bibitem{kuhn2023semantic}
Lorenz Kuhn, Yarin Gal, and Sebastian Farquhar.
\newblock Semantic uncertainty: Linguistic invariances for uncertainty estimation in natural language generation.
\newblock {\em arXiv preprint arXiv:2302.09664}, 2023.

\bibitem{roth2022uncertain}
Aaron Roth.
\newblock Uncertain: Modern topics in uncertainty estimation, 2022.

\bibitem{platt1999probabilistic}
John Platt et~al.
\newblock Probabilistic outputs for support vector machines and comparisons to regularized likelihood methods.
\newblock {\em Advances in large margin classifiers}, 10(3):61--74, 1999.

\bibitem{zadrozny2002transforming}
Bianca Zadrozny and Charles Elkan.
\newblock Transforming classifier scores into accurate multiclass probability estimates.
\newblock In {\em Proceedings of the eighth ACM SIGKDD international conference on Knowledge discovery and data mining}, pages 694--699, 2002.

\bibitem{zadrozny2001obtaining}
Bianca Zadrozny and Charles Elkan.
\newblock Obtaining calibrated probability estimates from decision trees and naive bayesian classifiers.
\newblock In {\em Icml}, volume~1, pages 609--616, 2001.

\bibitem{chen2016xgboost}
Tianqi Chen and Carlos Guestrin.
\newblock Xgboost: A scalable tree boosting system.
\newblock In {\em Proceedings of the 22nd acm sigkdd international conference on knowledge discovery and data mining}, pages 785--794, 2016.

\bibitem{breiman2001random}
Leo Breiman.
\newblock Random forests.
\newblock {\em Machine learning}, 45:5--32, 2001.

\bibitem{joshi2017triviaqa}
Mandar Joshi, Eunsol Choi, Daniel~S Weld, and Luke Zettlemoyer.
\newblock Triviaqa: A large scale distantly supervised challenge dataset for reading comprehension.
\newblock {\em arXiv preprint arXiv:1705.03551}, 2017.

\bibitem{wei2021finetuned}
Jason Wei, Maarten Bosma, Vincent~Y Zhao, Kelvin Guu, Adams~Wei Yu, Brian Lester, Nan Du, Andrew~M Dai, and Quoc~V Le.
\newblock Finetuned language models are zero-shot learners.
\newblock {\em arXiv preprint arXiv:2109.01652}, 2021.

\bibitem{jiang2023mistral}
Albert~Q Jiang, Alexandre Sablayrolles, Arthur Mensch, Chris Bamford, Devendra~Singh Chaplot, Diego de~las Casas, Florian Bressand, Gianna Lengyel, Guillaume Lample, Lucile Saulnier, et~al.
\newblock Mistral 7b.
\newblock {\em arXiv preprint arXiv:2310.06825}, 2023.

\bibitem{thorne2018fever}
James Thorne, Andreas Vlachos, Christos Christodoulopoulos, and Arpit Mittal.
\newblock Fever: a large-scale dataset for fact extraction and verification.
\newblock {\em arXiv preprint arXiv:1803.05355}, 2018.

\bibitem{li2023halueval}
Junyi Li, Xiaoxue Cheng, Wayne~Xin Zhao, Jian-Yun Nie, and Ji-Rong Wen.
\newblock Halueval: A large-scale hallucination evaluation benchmark for large language models.
\newblock In {\em Proceedings of the 2023 Conference on Empirical Methods in Natural Language Processing}, pages 6449--6464, 2023.

\bibitem{srivastava2022beyond}
Aarohi Srivastava, Abhinav Rastogi, Abhishek Rao, Abu Awal~Md Shoeb, Abubakar Abid, Adam Fisch, Adam~R Brown, Adam Santoro, Aditya Gupta, Adri{\`a} Garriga-Alonso, et~al.
\newblock Beyond the imitation game: Quantifying and extrapolating the capabilities of language models.
\newblock {\em arXiv preprint arXiv:2206.04615}, 2022.

\bibitem{brier1950verification}
Glenn~W Brier.
\newblock Verification of forecasts expressed in terms of probability.
\newblock {\em Monthly weather review}, 78(1):1--3, 1950.

\bibitem{jiang2024mixtral}
Albert~Q Jiang, Alexandre Sablayrolles, Antoine Roux, Arthur Mensch, Blanche Savary, Chris Bamford, Devendra~Singh Chaplot, Diego de~las Casas, Emma~Bou Hanna, Florian Bressand, et~al.
\newblock Mixtral of experts.
\newblock {\em arXiv preprint arXiv:2401.04088}, 2024.

\bibitem{falcon40b}
Ebtesam Almazrouei, Hamza Alobeidli, Abdulaziz Alshamsi, Alessandro Cappelli, Ruxandra Cojocaru, Merouane Debbah, Etienne Goffinet, Daniel Heslow, Julien Launay, Quentin Malartic, Badreddine Noune, Baptiste Pannier, and Guilherme Penedo.
\newblock {Falcon-40B}: an open large language model with state-of-the-art performance.
\newblock 2023.

\bibitem{together2023redpajama}
Together Computer.
\newblock Redpajama-data: An open source recipe to reproduce llama training dataset, 2023.

\bibitem{touvron2023llama}
Hugo Touvron, Thibaut Lavril, Gautier Izacard, Xavier Martinet, Marie-Anne Lachaux, Timoth{\'e}e Lacroix, Baptiste Rozi{\`e}re, Naman Goyal, Eric Hambro, Faisal Azhar, et~al.
\newblock Llama: Open and efficient foundation language models.
\newblock {\em arXiv preprint arXiv:2302.13971}, 2023.

\bibitem{openlm2023openllama}
Xinyang Geng and Hao Liu.
\newblock Openllama: An open reproduction of llama, May 2023.

\bibitem{wolf2019huggingface}
Thomas Wolf, Lysandre Debut, Victor Sanh, Julien Chaumond, Clement Delangue, Anthony Moi, Pierric Cistac, Tim Rault, R{\'e}mi Louf, Morgan Funtowicz, et~al.
\newblock Huggingface's transformers: State-of-the-art natural language processing.
\newblock {\em arXiv preprint arXiv:1910.03771}, 2019.

\bibitem{detommaso2023fortuna}
Gianluca Detommaso, Alberto Gasparin, Michele Donini, Matthias Seeger, Andrew~Gordon Wilson, and Cedric Archambeau.
\newblock Fortuna: A library for uncertainty quantification in deep learning.
\newblock {\em arXiv preprint arXiv:2302.04019}, 2023.

\bibitem{li2023angle}
Xianming Li and Jing Li.
\newblock Angle-optimized text embeddings.
\newblock {\em arXiv preprint arXiv:2309.12871}, 2023.

\bibitem{muennighoff2022mteb}
Niklas Muennighoff, Nouamane Tazi, Lo{\"\i}c Magne, and Nils Reimers.
\newblock Mteb: Massive text embedding benchmark.
\newblock {\em arXiv preprint arXiv:2210.07316}, 2022.

\bibitem{mcinnes2018umap-software}
Leland McInnes, John Healy, Nathaniel Saul, and Lukas Grossberger.
\newblock Umap: Uniform manifold approximation and projection.
\newblock {\em The Journal of Open Source Software}, 3(29):861, 2018.

\bibitem{schwarz1978estimating}
Gideon Schwarz.
\newblock Estimating the dimension of a model.
\newblock {\em The annals of statistics}, pages 461--464, 1978.

\bibitem{liu2024exploring}
Fang Liu, Yang Liu, Lin Shi, Houkun Huang, Ruifeng Wang, Zhen Yang, and Li~Zhang.
\newblock Exploring and evaluating hallucinations in llm-powered code generation.
\newblock {\em arXiv preprint arXiv:2404.00971}, 2024.

\bibitem{kalai2023calibrated}
Adam~Tauman Kalai and Santosh~S Vempala.
\newblock Calibrated language models must hallucinate.
\newblock {\em arXiv preprint arXiv:2311.14648}, 2023.

\end{thebibliography}

\appendix

\section{Additional Results}
\label{apdx:add}

\begin{table*}
\centering
\caption{Hallucination detection results on all datasets of scoring methods \textbf{without} calibration. Methods that are not applicable for the given dataset are marked as ---.}
\label{tab:res_nocal}
\resizebox{\textwidth}{!}{
\begin{tabular}{@{}llllllllllllllll@{}}
\toprule[1pt]
                        & \multicolumn{3}{l}{\textbf{TriviaQA}} & \multicolumn{3}{l}{\textbf{HaluEval}} & \multicolumn{3}{l}{\textbf{BIG-Bench}} & \multicolumn{3}{l}{\textbf{FEVER}} \\ \midrule
\textbf{Scoring Method} & Brier  & F1     & Acc    & Brier  & F1     & Acc    & Brier  & F1     & Acc    & Brier & F1 & Acc    \\ \midrule
\midrule
P(True)                 & 0.2467 & 0.8267 & 0.7487 & 0.3850 & 0.7195 & 0.6110 & 0.3631 & 0.4981 & 0.6117 & 0.0738 & 0.9241 & 0.9248  \\
P(True) Verbalized      & 0.2333 & 0.8056 & 0.7269 & 0.3957 & 0.6983 & 0.5645 & 0.3532 & 0.5531 & 0.5245 & 0.0807 & 0.9166 & 0.9163  \\
P(InputContradict)      & 0.2526 & 0.8129 & 0.7392 & 0.4045 & 0.7103 & 0.5910 & 0.3603 & 0.4469 & 0.6001 & 0.0578 & 0.9406 & 0.9398  \\
P(SelfContradict)       & 0.6438 & 0.1668 & 0.3402 & 0.4825 & 0.6635 & 0.5105 & 0.4238 & 0.2332 & 0.5257 & 0.2101 & 0.8037 & 0.7801  \\
P(FactContradict)       & 0.5571 & 0.3653 & 0.4152 & 0.4634 & 0.6708 & 0.5260 & 0.4076 & 0.2066 & 0.5389 & 0.1952 & 0.8100 & 0.7972  \\
\midrule
Inverse Perplexity      & 0.5358 & 0.1286 & 0.3541 & 0.0118 & 0.4980 & 0.3090 & 0.0784 & 0.1771 & 0.6353 & 0.4381 & 0.0055 & 0.4847 \\
\midrule
NLI (DeBERTa)           & 0.3666 & 0.5914 & 0.5102 & 0.4676 & 0.4910 & 0.2423 & 0.0086 & 0.1943 & 0.5413 & 0.2493 & 0.5987 & 0.5175  \\
\midrule
SelfCheckGPT-NLI        & 0.1503 & 0.8654 & 0.8022 & ---   & ---   & ---   & ---   & ---   & ---   & ---  & ---   & --- \\
HallucinationRail       & 0.2698 & 0.8240 & 0.7086 & ---   & ---   & ---   & ---   & ---   & ---   & ---  & ---   & --- \\
SimilarityDegree        & 0.1854 & 0.8488 & 0.7612 & ---   & ---   & ---   & ---   & ---   & ---   & ---  & ---   & --- \\
\bottomrule[1pt]
\end{tabular}%
}
\end{table*}

\begin{table*}
\centering
\caption{Hallucination detection results on all datasets of scoring methods after calibration via multicalibration. Methods that are not applicable for the given dataset are marked as ---.}
\label{tab:res_mistral_7b}
\resizebox{\textwidth}{!}{
\begin{tabular}{@{}llllllllllllllll@{}}
\toprule[1pt]
                        & \multicolumn{3}{l}{\textbf{TriviaQA}} & \multicolumn{3}{l}{\textbf{HaluEval}} & \multicolumn{3}{l}{\textbf{BIG-Bench}} & \multicolumn{3}{l}{\textbf{FEVER}} \\ \midrule
\textbf{Scoring Method} & Brier  & F1     & Acc    & Brier  & F1     & Acc    & Brier  & F1     & Acc    & Brier & F1 & Acc    \\ \midrule
\midrule
\texttt{Mixtral-8x7B-Instruct-v0.1}                &       &           &       &       &           &        &          &       &       &       &       &           &       &       \\
\midrule
P(True)                 & 0.1543 & 0.8695 & 0.8089 & 0.1959 & 0.7618 & 0.6970 & 0.2081 & 0.6347 & 0.6399 & 0.0769 & 0.9143 & 0.9145  \\
P(True) Verbalized      & 0.1499 & 0.8617 & 0.7983 & 0.2235 & 0.7158 & 0.6220 & 0.2177 & 0.5907 & 0.6290 & 0.0682 & 0.9221 & 0.9212  \\
P(InputContradict)      & 0.1772 & 0.8493 & 0.7687 & 0.2011 & 0.7547 & 0.6795 & 0.2177 & 0.5907 & 0.6290 & 0.0665 & 0.9288 & 0.9277  \\
P(SelfContradict)       & 0.2136 & 0.8151 & 0.6879 & 0.2458 & 0.6613 & 0.5550 & 0.2425 & 0.0000 & 0.6145 & 0.1603 & 0.8187 & 0.7979  \\
P(FactContradict)       & 0.2192 & 0.8151 & 0.6879 & 0.2387 & 0.6487 & 0.6030 & 0.2426 & 0.2923 & 0.5586 & 0.1470 & 0.8196 & 0.8190  \\
Inverse Perplexity      & 0.1894 & 0.8035 & 0.7175 & 0.2497 & 0.0672 & 0.5005 & 0.2455 & 0.2747 & 0.6192 & 0.2355 & 0.5188 & 0.5841 \\
\midrule
\texttt{OpenLLaMA 13B}                &       &           &       &       &           &        &          &       &       &       &       &           &       &       \\
\midrule
P(True)                 & 0.2133 & 0.8151 & 0.6879 & 0.2502 & 0.6640 & 0.5050 & 0.2495 & 0.0655 & 0.6047 & 0.2512 & 0.6402 & 0.5121  \\
P(True) Verbalized      & 0.2217 & 0.8151 & 0.6879 & 0.2502 & 0.6707 & 0.5045 & 0.2359 & 0.0620 & 0.6157 & 0.2486 & 0.6814 & 0.5167  \\
P(InputContradict)      & 0.2111 & 0.8150 & 0.6882 & 0.2502 & 0.6707 & 0.5045 & 0.2432 & 0.1730 & 0.5972 & 0.2500 & 0.6358 & 0.5207  \\
P(SelfContradict)       & 0.2179 & 0.8146 & 0.6874 & 0.2473 & 0.6743 & 0.5300 & 0.2440 & 0.4357 & 0.6099 & 0.2492 & 0.6547 & 0.4986  \\
P(FactContradict)       & 0.2218 & 0.8151 & 0.6879 & 0.2490 & 0.5972 & 0.5070 & 0.2414 & 0.2766 & 0.5926 & 0.2478 & 0.5577 & 0.5381  \\
Inverse Perplexity      & 0.2191 & 0.8151 & 0.6879 & 0.2466 & 0.2932 & 0.5300 & 0.2400 & 0.0933 & 0.6076 & 0.2404 & 0.4877 & 0.5688 \\
\midrule
\texttt{Falcon-7B-Instruct}                &       &           &       &       &           &        &          &       &       &       &       &           &       &       \\
\midrule
P(True)                 & 0.2128 & 0.8147 & 0.6874 & 0.2487 & 0.2794 & 0.5305 & 0.2395 & 0.0030 & 0.6145 & 0.2478 & 0.6198 & 0.5527  \\
P(True) Verbalized      & 0.2171 & 0.8151 & 0.6879 & 0.2509 & 0.6709 & 0.5055 & 0.2455 & 0.0030 & 0.6151 & 0.2505 & 0.6782 & 0.5153  \\
P(InputContradict)      & 0.2169 & 0.8127 & 0.6857 & 0.2475 & 0.3852 & 0.5515 & 0.2413 & 0.0000 & 0.6145 & 0.2493 & 0.5705 & 0.5203  \\
P(SelfContradict)       & 0.2138 & 0.8072 & 0.6826 & 0.2493 & 0.5307 & 0.5330 & 0.2347 & 0.0000 & 0.6145 & 0.2364 & 0.6897 & 0.6091  \\
P(FactContradict)       & 0.2135 & 0.8151 & 0.6879 & 0.2265 & 0.5802 & 0.6230 & 0.2364 & 0.0000 & 0.6145 & 0.2358 & 0.6521 & 0.6030  \\
Inverse Perplexity      & 0.2160 & 0.8151 & 0.6879 & 0.2491 & 0.3705 & 0.5260 & 0.2497 & 0.0809 & 0.5938 & 0.2408 & 0.4853 & 0.5752 \\
\bottomrule[1pt]
\end{tabular}%
}
\end{table*}

\end{document}